%% file: acl_latex.tex
\pdfoutput=1

\documentclass[11pt]{article}

\usepackage[final]{acl}
\usepackage{pgfplots}
\usepackage{times}
\usepackage{latexsym}
\usepackage{listings}
\usepackage{multirow}
\usepackage[T1]{fontenc}

\usepackage[utf8]{inputenc}

\usepackage{microtype}

\usepackage{inconsolata}
\usepackage{mathabx}

\usepackage{scalerel,stackengine}

\usepackage{graphicx}
\usepackage{ifthen}
\usepackage{csvsimple}
\usepackage{pgffor}
\usepackage{pgfplots}
\usepackage{pgfplotstable}
\usepackage{subcaption}
\usepackage{caption}
\usepgfplotslibrary{colorbrewer}
\usetikzlibrary{pgfplots.colorbrewer}
\pgfplotsset{compat=1.18, cycle list/Set3,}

\pgfplotsset{
    discard if not/.style 2 args={
      x filter/.append code={
        \edef\tempa{\thisrow{#1}}
        \edef\tempaa{#2}
        \ifx\tempa\tempaa
        \else
            
        \fi
      }
    }}

    \usepackage{url}

\DeclareUrlCommand\UScore{\urlstyle{rm}}
\newcommand{\expUScore}{%
  \expandafter\expandafter\expandafter
  \UScore
  \expandafter\expandafter\expandafter
}

\title{Acquiescence Bias in Large Language Models}

\author{Daniel Braun \\
 Marburg University\\
  Department of Mathematics and Computer Science\\
  \texttt{daniel.braun@uni-marburg.de}}

\begin{document}
\maketitle
\begin{abstract}
Acquiescence bias, i.e. the tendency of humans to agree with statements in surveys, independent of their actual beliefs, is well researched and documented. Since Large Language Models (LLMs) have been shown to be very influenceable by relatively small changes in input and are trained on human-generated data, it is reasonable to assume that they could show a similar tendency. We present a study investigating the presence of acquiescence bias in LLMs across different models, tasks, and languages (English, German, and Polish). Our results indicate that, contrary to humans, LLMs display a bias towards answering \textit{no}, regardless of whether it indicates agreement or disagreement.
\end{abstract}

\section{Introduction}
\label{sec:intro}
When humans are asked whether they agree with something, they have a bias towards doing so, independent of their actual beliefs. In other words, the question ``Is the weather good or bad?'' can lead to different responses than asking ``Is the weather good?'' The phenomenon, called acquiescence bias, is well documented and has been studied for decades (see Section \ref{sec:relwork}). One guideline for good survey design is, therefore, to avoid such yes/no or agree/disagree questions.

Large Language Models (LLMs) have shown to be sensitive towards prompt variations \citep{zhuo-etal-2024-prosa,anagnostidis2024susceptible,loya-etal-2023-exploring,rottger-etal-2024-political,halleryes} and to replicate a variety of cognitive biases found in humans, e.g. with regard to anchoring \cite{jones2022capturing}, item order \citep{koo-etal-2024-benchmarking}, and group attribution \citep{echterhoff-etal-2024-cognitive}. It is therefore somewhat reasonable to assume that LLMs could also display some form of acquiescence bias, which could have implications on how prompts should be designed and whether or not LLMs can be used to simulate human responses. In this paper, we evaluate five LLMs of different sizes (namely Llama-3.1-8B-Instruct, Mistral-Small-24B-Instruct-2501, gemma-2-27b-it, Llama-3.3-70B-Instruct, and gpt-4o-2024-08-06) on nine different tasks in three languages (English, German, and Polish), to investigate whether they show response patterns resembling acquiescence bias. Our results indicate that changing questions into a yes/no format has significant influence on the responses of LLMs, across tasks and models, and the tested languages. While no clear pattern emerged for German and Polish, for English, we found that, unlike humans, LLMs are biased towards answering \textit{no}, independent of whether that indicates agreement or disagreement. These findings do not only imply that the design of prompts should be considered very carefully but also that LLMs are not well suited to simulate human responses to survey questions.

\section{Related Work}
\label{sec:relwork}

Acquiescence bias, i.e. ``the tendency to endorse any assertion made in a question,
regardless of its content'' \cite{annurev:/content/journals/10.1146/annurev.psych.50.1.537}, has been found in surveys across multiple domains, from political questions \citep{wright1975does, hill2023acquiescence} to assessing personalities \citep{rammstedt2013impact, danner2015acquiescence}. Possible explanations for this bias range from the tendency to present socially acceptable behaviour by not disagreeing with an assumed authority of the questionnaire (see \citet{hill2023acquiescence} for a more detailed discussion). According to \citet{annurev:/content/journals/10.1146/annurev.psych.50.1.537}, the size of the effect is estimated to be around 10\%.

While researchers have extensively investigated a variety of cognitive biases found in humans and whether LLMs replicate them \citep{jones2022capturing,koo-etal-2024-benchmarking,echterhoff-etal-2024-cognitive}, \citet{tjuatja2024llms} were the first to explicitly investigate whether LLMs replicate acquiescence bias, alongside with four other human biases relevant in survey design. While they found that LLMs are indeed sensitive to changes in prompts, they did not find consistent patterns in these changes. Despite these initial results, we believed, a deeper analysis of the acquiescence bias in particular was warranted. Given the broader scope of the work by \citet{tjuatja2024llms}, only 176 questions and 352 responses have been evaluated with regard to acquiescence bias. In this paper, we analyse responses to more than 37,975 question variations. Additionally, we address a limitation identified by \citet{tjuatja2024llms}, by not only using English corpora but also other languages, namely German and Polish\footnote{The additional languages have been chosen based on the availability of appropriate datasets and the ability of the authors to interpret the results.}. Lastly, we believed that the way in which the questions were adapted by \citeauthor{tjuatja2024llms} could have potentially added noise that influenced the results. In order to turn the original questions into yes/no questions they added the phrases ``don't you agree'' or ``wouldn't you agree'' to all questions (e.g. ``On social media, do you think of yourself more as a A. Sharer of news B. Receiver of news'' was turned into ``On social media, wouldn't you say you are more of a sharer of news than a receiver of news? A. Yes B. No''). Such questions are also called \textit{negative yes/no questions} and carry inherent ambiguity, it is not always clear whether answering them with ``yes'' indicates agreement or disagreement \citep{romero2004negative}.
Additionally, negative modifier (\textit{wouldn't} you agree instead of \textit{would} you agree) also have an influence on human responses \citep{johnson2004did}. Therefore, we believe they are not well suited to investigate a potential acquiescence bias.

\section{Experimental Design}
\subsection{Data}
To investigate whether LLMs show response patterns similar to the acquiescence bias observed in humans, we used binary questions from nine different tasks. Seven of those tasks are English and stem from the Legelbench corpus \citep{guha2024legalbench}, one is German and stems from the AGB-DE corpus \citep{braun-matthes-2024-agb}, and one is Polish and stems from the LEPISZCZE corpus \citep{NEURIPS2022_890b206e}. While the tasks chosen from the Legalbench dataset are originally phrased as yes/no questions, the other two datasets are originally designed to be A/B choices (valid/void or abusive/safe). The tasks were chosen based on how well they could be transferred from A/B to yes/no question, in order to avoid as much rephrasing as possible, in an attempt to isolate a potential acquiescence bias as much as possible from other influences.

Although all tasks are performed on legal documents, not all of them are inherently legal. Out of the nine tasks, three are linguistic in nature: two are concerned with Natural Language Inference (NLI) and one with the classification of definitions. For the English language data, these three linguistic tasks represent 45\% of the overall data. Of the remaining six tasks, two are concerned with finding void clauses in contracts, one with classifying whether a statement constitute hearsay, and three with legal clause classification tasks. A detailed description of each tasks and the distribution of the data can be found in Appendix \ref{sec:taskdescdetail}.

Nevertheless, the fact remains that all texts are from the legal domain. Although legal texts are increasingly used as benchmark for the capabilities of LLMs (see e.g. \citet{fei-etal-2024-lawbench} and \citet{steging2025parameterizedargumentationbasedreasoningtasks}), it has also been shown that they provide specific challenges to LLMs which are different from texts from other domains \citep{jayakumar-etal-2023-large}. Therefore, when interpreting the results of this study and how far they can be generalized, the composition of the investigated data should be kept in mind.

\subsection{Prompts}
In order to evaluate the influence that the phrasing of prompts has on the answers generated by LLMs, for each of the questions in the datasets, we tested five different ways to phrase them :

\begin{itemize}
    \item Neutral: two options are presented (e.g. <<clause text>> What kind of clause is this? Just answer ``Valid'' or ``Void''.).
    \item Yes/No: the first option is presented as a yes/no question (e.g. <<clause text>> Is this clause valid? Just answer ``Yes'' or ``No''.). A ``yes'' response here is equivalent to choosing option A. An increase from A to yes could therefore indicate acquiescence bias.
    \item Agreement: the first option is presented as response and the LLM is asked to agree (e.g. <<clause text>> Do you agree that this clause is valid? Just answer ``Yes'' or ``No''.). Similarly, a ``yes'' response is indicating response A here, however the agreement aspect is emphasized. In case of an acquiescence bias we expect more ``yes'' answers than ``A'' answers before.
    \item Negated agreement: same as agreement but instead of ``do you agree'' the phrasing ``don't you agree'' is used (e.g. <<clause text>> Don't you agree that this clause is valid? Just answer ``Yes'' or ``No''.). As mentioned, this condition is ambiguous, therefore it is not necessarily clear how an acquiescence bias would influence responses to this type of question.
    \item Disagreement: the first option is presented as response and the LLM is asked to disagree (e.g. <<clause text>> Do you disagree that this clause is valid? Just answer ``Yes'' or ``No''.). In this condition, ``no'', due to the double negation, implies agreement with the question (and therefore option A).
\end{itemize}

The different conditions were designed to be as similar as possible, in order to ensure that any changes in the response are only caused by the changing level of agreement that is suggested by the phrasing. E.g. the first element for the neutral question was always the later ``yes'' answer, except for the disagreement prompt, to avoid biases introduced by the order of items.
All prompts have been designed to instruct the models to reply with just one word. In this way, we aim to replicate survey style situations in which acquiescence bias is most often studied in humans and simplify the processing of responses (see Section \ref{sec:responseprocessing}). 

\input{overview_table}

\subsection{Hardware}
The experiments with the open weight models (Llama, Mistral, and Gemma) were conducted on an HPC cluster using four Nvidia A100 GPUs. The total compute time for generating all 152,040 responses (38,010 variations for each of the four models) was 10 hours and 40 minutes. For the GPT-4 model, the OpenAI API was used. The total costs were \$12.38. All models were used with standard parameters (see Appendix \ref{sec:hyperparameters} for details), including a temperature of 1.0, because we believe standard parameters have the highest practical relevance and, as shown by \citet{renze-2024-effect}, ``changes in temperature from 0.0 to 1.0 do not have a statistically significant impact on LLM performance''. The code that was used for the evaluation, as well as the responses retrieved are published on GitHub under the MIT license\footnote{\url{https://github.com/Responsible-NLP/Acquiescence-Bias-in-Large-Language-Models}}. Each request was completely independent of each other to avoid an influence of previous responses.

\subsection{Response Processing}
\label{sec:responseprocessing}
Overall, all models followed the instruction to reply with only one word relatively well, thereby easing the processing of responses. Of the 38,010 A/B responses that we processed across all models and languages, 92.4\% only contained one of the options (as prompted) at most followed by a dot. 7.5\% contained one of the options at the beginning of the response, followed by an additional text (mostly explanation). Only less than 0.1\% (a total of 67) did not start with one of the options, all of those still contained one of the options in the answer, mostly at the very end after a disclaimer that the answer depends on additional context. Of the 152,788 processed yes/no responses, 99.4\% just contained yes or no (sometimes followed by a dot). 0.003\% contained yes or no at the beginning followed by a text, and only a total of 369 responses (0.002\%) did not contain a yes / no response at the beginning, but again mostly at the end of the response.

\section{Results}

The main metric we were interested in originally was the number of positives (the sum of true and false positives), i.e. answers that are equivalent to choosing option A. If LLMs indeed show response patterns equivalent to acquiescence bias in humans, we would expect to see an increase in positives (irrespective of whether they are true or false positives) when changing from the neutral prompt to the other options. For the disagreement prompt, responses are inverted because a ``yes'' here indicates disagreement which correlates with a ``no'' in the original data. Therefore, a higher positive rate here means that the model responded with ``no'' more often, which indicates agreement with the question and is equivalent to option A.

Across all languages and models, we did see that the different prompts have a significant influence on the responses. However, we did not find that models are systematically biased towards agreeing (detailed results can be found in Appendix \ref{sec:appendix}). On the contrary, we found that for most cases, the number of positives was significantly reduce in comparison to responses of ``A'' in the neutral prompt. What we found instead is a bias towards replying ``no'' independent of whether that indicates disagreement or agreement with the question\footnote{I.e. we also did not find the opposite of an acquiescence bias, a tendency towards disagreeing, but simply a tendency to reply \textit{no}.}. 

Instead of focusing on positives, we decided to focus the evaluation on the absolute number of responses, particularly ``no'' responses. Table \ref{tab:abs} shows an overview of the change in responses. In English, changing the neutral A/B question to a yes/no question increased the number of responses that are equivalent to B between 31\% and 203\% across all models. A similar effect can be seen for the agree yes/no question. As initially suspected, adding a negative pre-modifier (``don't you agree'' instead of ``do you agree'') indeed changes the outcome and decreases the effect size, yet we still see a consistent increase in B/no answers. This is different from the results reported by \citet{tjuatja2024llms}, where no clear pattern was visible when the neutral questions were just transformed to negative yes/no questions. However, it is worth pointing out these results are based on the previous generation of models (i.e. Llama2 and GPT-3).

Most surprisingly was that we see the same pattern for all but one model for the disagree questions in which the increase of no is still consistent, despite the fact that it is logically contradictory to the previous conditions. The results indicate that, for English, the models are not biased towards disagreeing, like humans are biased towards agreeing, but simply are biased towards replying ``no'', independent of whether that indicates agreement or disagreement. To validate our findings, we conducted McNemar's tests (within-subjects chi-squared test,  \citet{mcnemar1947note}) for all tasks individually and the overall results. The results of the analysis (see Appendix \ref{sec:appendix}) show that all conditions significantly influenced the answers overall. Even on the individual task level, despite the much smaller number of samples, almost all effects are statistically significant at the threshold of $p < 0.05$.
While in German and Polish the phrasing of the prompt also has a significant influence on the results, no clear pattern could be identified across models or conditions. 

With regard to accuracy, we see that the influence that the observed bias has on accuracy is heavily based on the type of errors that are most prevalent in the neutral setting. For the Lllama-3.3-70B model on the cuad\_non-compete dataset, for example, there is a very small number of false positives in the neutral setting (total of 9 or 2\%). Changing to a condition which has a bias towards no (like the agree prompt), in such cases decreases accuracy. In other settings, where there is a higher number of false positives, e.g. Mistral-Small-24B on the contract\_nli\_confidentiality\_of\_agreement datasets, with a total of 40 false positives (48\%), changing to a condition with a bias towards no increases accuracy.

Table \ref{tab:change} shows how the amount of true and false negatives changes in percent for all English tasks, compared to the neutral condition. Although true negatives often increase more strongly than false negatives, thereby improving accuracy, this is not always the case. This is in line with our observation that the models are biased towards the response “no”, independent of the logical implication and thereby also the implication on overall accuracy.

\input{change_table}

\section{Conclusion}

Our initial hypothesis, that LLMs could display a bias resembling the acquiescence bias, was not confirmed. While we did find a systematic bias that could be observed across models and tasks (although only in English), interestingly it was also not the opposite, i.e. a bias towards disagreeing, but a bias towards replying ``no'', independent of whether that indicates agreement or disagreement. 

We believe that these findings bear importance, not only because they indicate that LLMs do not replicate response biases found in humans making it questionable whether they can be used to simulate human responses, but also because they shed doubt on the reasoning abilities shown by these models. If the responses would be mainly based on reasoning, we would not expect to see significant differences in the responses between the questions ``Do you agree sentence X is a definition?'', and ``Do you disagree sentence X is a definition?''. However, we found that in both cases models, irrespective of their size, are biased towards answering ``no'', thereby contradicting previous responses.

\section*{Limitations}

The presented study has limitations with regard to the generalisability of its results:

\begin{itemize}
    \item While we did investigate models of different sizes and found consistent patterns across those models, it is unclear if they generalise beyond them.
    \item While our results show significant influence of the phrasing of questions across all three investigated languages, a consistent pattern that is representative of a bias could only be identified in English.
    \item All datasets that were used stem from the legal domain. Although not all tasks are inherently of legal nature (e.g. assessing whether a sentence is a definition or not), further investigation is necessary to find out whether the findings are also applicable to documents from other domains.
    \item Only one prompt was tested per condition. The phrasing of the prompts can have significant influence on the results and other prompts could have been found that fit the described conditions. The prompts were chosen to be as concise as possible in order to minimize changes not related to the conditions themselves. 
    \item LLM outputs contain an inherent degree of randomness and re-running the same prompts would most likely lead to slightly different results. However given the large number of analysed responses, the consistent pattern, and the statistical significance of our results, we are confident that the observed variations are not just random noise. 
\end{itemize}

\bibliography{custom}

\appendix
\section{Detailed Task Description}
\label{sec:taskdescdetail}

Table \ref{tab:tasks} shows the number of questions for each of the tasks that have been used.

\begin{table*}
    \centering
      \caption{Corpora and tasks used in this study \label{tbl:tasks}}
    \label{tab:tasks}
    \begin{tabular}{l l l r}
    \hline
       \textbf{Corpus} & \textbf{Task} & \textbf{Lang.} & \textbf{\# questions} \\\hline
      Legelbench & hearsay & en & 93\\
      Legelbench & definition\_classification & en & 1,336\\
      Legelbench & cuad\_non-compete & en & 441\\
      Legelbench & cuad\_no-solicit\_of\_customers & en & 83\\
      Legelbench & cuad\_cap\_on\_liability& en & 1,245\\
      Legelbench & contract\_nli\_explicit\_identification & en & 108\\
      Legelbench & contract\_nli\_confidentiality\_of\_agreement & en & 81\\
      AGB-DE & agb-de & de & 755\\
      LEPISZCZE & clauses-pl & pl & 3,453\\ \hline
      $\sum$ & & & 7,595\\\hline
    \end{tabular}

\end{table*}

\subsection{Legalbench}
\begin{itemize}
    \item hearsay: ``We create a dataset to test a model’s ability to apply the hearsay rule. Each sample in the dataset describes (1) an issue being litigated or an assertion a party wishes to prove, and (2) a piece of evidence a party wishes to introduce. The goal is to determine if—as it relates to the issue—the evidence would be considered hearsay [...].''\footnote{\url{https://hazyresearch.stanford.edu/legalbench/tasks/hearsay.html}, last accessed 16.05.2025}
    \item definition\_classification: ``The goal of this task is to identify if a sentence contains a definition. For example, the following sentence defines “vacation”: \texttt{A vacation is defined by Bouvier to be the period of time between the end of one term and the beginning of another.}''\footnote{\url{https://hazyresearch.stanford.edu/legalbench/tasks/definition_classification.html}, last accessed 16.05.2025}
    \item cuad\_non-compete: ``This is a binary classification task in which the model must determine if a contractual clause falls under the category of “Non-Compete”.''\footnote{\url{https://hazyresearch.stanford.edu/legalbench/tasks/cuad_non-compete.html}, last accessed 16.05.2025}
    \item cuad\_no-solicit\_of\_customers: ``This is a binary classification task in which the model must determine if a contractual clause falls under the category of “No-Solicit Of Customers”.''\footnote{\url{https://hazyresearch.stanford.edu/legalbench/tasks/cuad_no-solicit_of_customers.html}, last accessed 16.05.2025}
    \item cuad\_cap\_on\_liability: ``This is a binary classification task in which the model must determine if a contractual clause falls under the category of “Cap On Liability”.'' \footnote{\url{https://hazyresearch.stanford.edu/legalbench/tasks/cuad_cap_on_liability.html}, last accessed 16.05.2025}
    \item contract\_nli\_explicit\_identification: ``This task was constructed from the ContractNLI dataset, which originally annotated clauses from NDAs based on whether they entailed, contradicted, or neglgected to mention a hypothesis. We binarized this dataset, treating contradictions and failures to mention as the negative label. We used the hypothesis provided as the prompt. Please see the original paper for more information on construction. All samples are drawn from the test set.'' \footnote{\url{https://hazyresearch.stanford.edu/legalbench/tasks/contract_nli_explicit_identification.html}, last accessed 16.05.2025}
    \item contract\_nli\_confidentiality\_of\_agreement: ``This task was constructed from the ContractNLI dataset, which originally annotated clauses from NDAs based on whether they entailed, contradicted, or neglgected to mention a hypothesis. We binarized this dataset, treating contradictions and failures to mention as the negative label. We used the hypothesis provided as the prompt. Please see the original paper for more information on construction. All samples are drawn from the test set.''\footnote{\url{https://hazyresearch.stanford.edu/legalbench/tasks/contract_nli_confidentiality_of_agreement.html}, last accessed 16.05.2025}
\end{itemize}

\subsection{AGB-DE}
``A clause in a contract is void, i.e. cannot be enforced by the parties of the contract,
if it contradicts governing law. Whether a clause is actually void depends on many things, including, in some cases, whether one of the parties is a consumer or whether both parties are businesses. The final decision on whether a specific clause in specific circumstances is actually void can only be made by a court of law. Therefore, the instruction for the annotators was to label a clause as potentially void, if they think a consumer residing in Germany could successfully challenge the clause in court.'' \cite{braun-matthes-2024-agb}

\subsection{LEPISZCZE}
``The PAC aims to detect “unfair contractual term” as the equivalent of
an abusive clause. The task was formulated as binary text classification. The dataset has been created with the Office of Competition and Consumer Protection.'' \cite{NEURIPS2022_890b206e}

\section{Hyperparameters}
\label{sec:hyperparameters}

The hyperparameters (temperature, top-p, and max\_new\_tokens) that have been used for the generation with the different models are shown in Table \ref{tab:hyper}.

\begin{table*}
    \caption{Hyperparameters used for text generation}
    \label{tab:hyper}
    \centering
    \begin{tabular}{l l l l}
    \hline
       \textbf{Model} & \textbf{temperature} & \textbf{top-p} & \textbf{max\_new\_tokens}  \\\hline
      Llama-3.1-8B & 1.0 & 1.0 & 1000\\
      Mistral-Small-24B & 1.0 & 1.0 & 1000\\
      gemma-2-27b-it & 1.0 & 1.0 & 1000\\
      Llama-3.3-70B & 1.0 & 1.0 & 1000\\
      gpt-4o-2024-08-0 & 1.0 & 1.0 & null\\\hline
    \end{tabular}

\end{table*}

\section{Prompts}
\label{sec:prmopts}

In this section, we introduce the prompts that have been used during the experiment. The same \texttt{system\_prompt} was used in all conditions.

\subsection{hearsay}
\begin{lstlisting}
{
    "name": "hearsay",
    "system_prompt": "Hearsay is an out-of-court statement introduced to prove the truth of the matter asserted. Determine if the evidence in the following fact patterns is hearsay under the Federal Rules of Evidence.",
    "neutral_prompt": "What is this statement?",
    "neutral_anwers": "Just answer \"Hearsay\" or \"Other Statement\".",
    "yesno_prompt": "Is there hearsay?",
    "agree_prompt": "Do you agree there is hearsay?",
    "negated_agree_prompt": "Don't you agree there is hearsay?",
    "disagree_prompt": "Do you disagree there is hearsay?"
}
\end{lstlisting}
\subsection{definition\_classification}
\begin{lstlisting}
{
    "name": "definition_classification",
    "neutral_prompt": "What type of sentence is this?",
    "neutral_anwers": "Just answer \"Definition\" or \"Other\".",
    "yesno_prompt": "Is the sentence defining a term?",
    "agree_prompt": "Do you agree that the sentence is defining a term?",
    "negated_agree_prompt": "Don't you agree that the sentence is defining a term?",
    "disagree_prompt": "Do you disagree that the sentence is defining a term?"
}
\end{lstlisting}
\subsection{cuad\_non-compete}
\begin{lstlisting}
{
    "name": "cuad_non-compete",
    "system_prompt": "A non-compete clause restricts the ability of a party to compete with the counterparty or operate in a certain geography or business or technology sector.",
    "neutral_prompt": "What kind of clause is that?",
    "neutral_anwers": "Just answer \"Non-Compete\" or \"Other\".",
    "yesno_prompt": "Does the clause restrict the ability of a party to compete with the counterparty or operate in a certain geography or business or technology sector?",
    "agree_prompt": "Do you agree that the clause restricts the ability of a party to compete with the counterparty or operate in a certain geography or business or technology sector?",
    "negated_agree_prompt": "Don't you agree that the clause restricts the ability of a party to compete with the counterparty or operate in a certain geography or business or technology sector?",
    "disagree_prompt": "Do you disagree that the clause restricts the ability of a party to compete with the counterparty or operate in a certain geography or business or technology sector?"
}
\end{lstlisting}
\subsection{cuad\_no-solicit\_of\_customers}
\begin{lstlisting}
{
    "name": "cuad_no-solicit_of_customers",
    "system_prompt": "Cap on liability upon the breach of a party's obligation includes time limitation for the counterparty to bring claims or maximum amount for recovery.",
    "neutral_prompt": "Does the clause restrict a party from contracting or soliciting customers or partners of the counterparty, whether during the contract or after the contract ends (or both)?",
    "neutral_anwers": "Just answer \"Restricted\" or \"Unrestricted\".",
    "yesno_prompt": "Does the clause restrict a party from contracting or soliciting customers or partners of the counterparty, whether during the contract or after the contract ends (or both)?",
    "agree_prompt": "Do you agree that the clause restricts a party from contracting or soliciting customers or partners of the counterparty, whether during the contract or after the contract ends (or both)?",
    "negated_agree_prompt": "Don't you agree that the clause restricts a party from contracting or soliciting customers or partners of the counterparty, whether during the contract or after the contract ends (or both)?",
    "disagree_prompt": "Do you disagree that the clause restricts a party from contracting or soliciting customers or partners of the counterparty, whether during the contract or after the contract ends (or both)?"
}
\end{lstlisting}
\subsection{cuad\_cap\_on\_liability}
\begin{lstlisting}
{
  "name": "cuad_cap_on_liability",
  "system_prompt": "Cap on liability upon the breach of a party's obligation includes time limitation for the counterparty to bring claims or maximum amount for recovery.",
  "neutral_prompt": "What does the clause specify with regard to the liability upon the breach of a party's obligation?",
  "neutral_anwers": "Just answer \"Cap\" or \"Uncapped\".",
  "yesno_prompt": "Does the clause specify a cap on liability upon the breach of a party's obligation?",
  "agree_prompt": "Do you agree that the clause specifies a cap on liability upon the breach of a party's obligation?",
  "negated_agree_prompt": "Don't you agree that the clause specifies a cap on liability upon the breach of a party's obligation?",
  "disagree_prompt": "Do you disagree that the clause specifies a cap on liability upon the breach of a party's obligation?"
}
\end{lstlisting}
\subsection{contract\_nli\_explicit\_identification}
\begin{lstlisting}
{
    "name": "contract_nli_explicit_ identification",
    "neutral_prompt": "What does the clause provide with regard to the identification of Confidential Information by the Disclosing Party?",
    "neutral_anwers": "Just answer \"Expressly\" or \"Other\".",
    "yesno_prompt": "Does the clause provide that all Confidential Information shall be expressly identified by the Disclosing Party?",
    "agree_prompt": "Do you agree that the clause provides that all Confidential Information shall be expressly identified by the Disclosing Party?",
    "negated_agree_prompt": "Don't you agree that the clause provides that all Confidential Information shall be expressly identified by the Disclosing Party?",
    "disagree_prompt": "Do you disagree that the clause provides that all Confidential Information shall be expressly identified by the Disclosing Party?"
}
\end{lstlisting}
\subsection{contract\_nli\_confidentiality\_of\_ agreement}
\begin{lstlisting}
{
    "name": "contract_nli_confidentiality_of_ agreement",
    "neutral_prompt": "What does the clause provide that the Receiving Party shall not do about the fact that Agreement was agreed or negotiated?",
    "neutral_anwers": "Just answer \"Disclose\" or \"Other\".",
    "yesno_prompt": "Does the clause provide that the Receiving Party shall not disclose the fact that Agreement was agreed or negotiated?",
    "agree_prompt": "Do you agree that the clause provides that the Receiving Party shall not disclose the fact that Agreement was agreed or negotiated?",
    "negated_agree_prompt": "Don't you agree that the clause provides that the Receiving Party shall not disclose the fact that Agreement was agreed or negotiated?",
    "disagree_prompt": "Do you disagree that the clause provides that the Receiving Party shall not disclose the fact that Agreement was agreed or negotiated?"
}
\end{lstlisting}
\subsection{AGB-DE}
\begin{lstlisting}
{
    "name": "agb-de",
    "system_prompt": "Stell dir vor, du bist ein Anwalt.",
    "neutral_prompt": "Wie würdest du diese Klausel bewerten?",
    "neutral_anwers": "Antworte nur \"ungültig\" oder \"gültig\".",
    "yesno_prompt": "Ist die Klausel ungültig?",
    "agree_prompt": "Stimmst du zu, dass die Klausel ungültig ist?",
    "negated_agree_prompt": "Stimmst du nicht zu, dass die Klausel ungültig ist?",
    "disagree_prompt": "Widersprichst du, dass die folgende Klausel ungültig ist?"
}
\end{lstlisting}
\subsection{LEPISZCZE}
\begin{lstlisting}
{
    "name": "clauses-pl",
    "neutral_prompt": "Jak oceniasz tę klauzulę?",
    "neutral_anwers": "Wystarczy odpowiedzieć \"Abuzywna\" lub \"Bezpieczne\".",
    "yesno_prompt": "Czy klauzula to jest nadużyciem?",
    "agree_prompt": "Czy zgadzasz się, że ta klauzula jest nadużyciem?",
    "negated_agree_prompt": "Czy nie zgadzasz się, że ta klauzula jest nadużyciem?",
    "disagree_prompt": "Czy zaprzecza Pan, że klauzula ta jest nadużyciem?"
    }
\end{lstlisting}

\section{Detailed Results}
\label{sec:appendix}

See Tables \ref{tab:Llama-3.1-8B} to \ref{tab:gpt-4o-2024-08-06}.

\input{table}

\end{document}

%% file: overview_table.tex
\begin{table*}
\small
    \centering
        \caption{Absolute response counts per model across languages and conditions (relative change for the \textit{no} option in comparison to the neutral condition in brackets)}
    \label{tab:abs}
    \begin{tabular}{l l l|l l l l l}\hline
        \textbf{Lang.} & \textbf{Condition} & \textbf{Response} & \textbf{Llama-3.1-8B} &  \textbf{Mistral-24B} & \textbf{Gemma} & \textbf{Llama-3.3-70B} &\textbf{GPT-4o}\\\hline
\multirow{10}{*}{DE}&\multirow{ 2}{*}{neutral}&A&444&592&413&143&132\\
&&B&311&163&342&612&623\\
&\multirow{ 2}{*}{yesno}&Yes (A)&34&233&372&159&283\\
&&No (B)&721 (132\%)&522 (220\%)&383 (12\%)&596 (-3\%)&472 (-24\%)\\
&\multirow{ 2}{*}{agree}&Yes (A)&83&155&607&331&234\\
&&No (B)&672 (116\%)&600 (268\%)&148 (-57\%)&424 (-31\%)&521 (-16\%)\\
&\multirow{ 2}{*}{negated}&Yes&36&129&338&458&205\\
&&No&719 (131\%)&626 (284\%)&417 (22\%)&297 (-51\%)&550 (-12\%)\\
&\multirow{ 2}{*}{disagree}&Yes (B)&642&490&270&680&520\\
&&No (A)&113 (-64\%)&265 (63\%)&485 (42\%)&75 (-88\%)&235 (-62\%)\\\cline{2-8}
\multirow{10}{*}{EN}&\multirow{ 2}{*}{neutral}&A&2915&2704&2985&1768&2129\\
&&B&479&690&409&1626&1265\\
&\multirow{ 2}{*}{yesno}&Yes (A)&2003&1306&2398&1263&1249\\
&&No (B)&1391 (190\%)&2088 (203\%)&996 (144\%)&2131 (31\%)&2145 (70\%)\\
&\multirow{ 2}{*}{agree}&Yes (A)&2279&1416&2557&1289&1417\\
&&No (B)&1115 (133\%)&1978 (187\%)&837 (105\%)&2105 (29\%)&1977 (56\%)\\
&\multirow{ 2}{*}{negated}&Yes&2532&1680&2703&1419&1604\\
&&No&862 (80\%)&1714 (148\%)&691 (69\%)&1975 (21\%)&1790 (42\%)\\
&\multirow{ 2}{*}{disagree}&Yes (B)&2004&2660&1789&3125&1918\\
&&No (A)&1390 (190\%)&734 (6\%)&1605 (292\%)&269 (-83\%)&1476 (17\%)\\\cline{2-8}
\multirow{10}{*}{PL}&\multirow{ 2}{*}{neutral}&A&1561&1583&3309&2465&1499\\
&&B&1892&1870&144&988&1954\\
&\multirow{ 2}{*}{yesno}&Yes (A)&2119&1069&2537&1668&1082\\
&&No (B)&1334 (-29\%)&2384 (27\%)&916 (536\%)&1785 (81\%)&2371 (21\%)\\
&\multirow{ 2}{*}{agree}&Yes (A)&2529&1050&2697&2452&1955\\
&&No (B)&924 (-51\%)&2403 (29\%)&756 (425\%)&1001 (1\%)&1498 (-23\%)\\
&\multirow{ 2}{*}{negated}&Yes&2350&2172&2985&2578&2160\\
&&No&1103 (-42\%)&1281 (-31\%)&468 (225\%)&875 (-11\%)&1293 (-34\%)\\
&\multirow{ 2}{*}{disagree}&Yes (B)&1864&2989&995&3145&2588\\
&&No (A)&1589 (-16\%)&464 (-75\%)&2458 (1607\%)&308 (-69\%)&865 (-56\%)\\\hline
    \end{tabular}

\end{table*}

%% file: change_table.tex
\begin{table}[]
\small
    \centering
        \caption{Change in true (TN) and false negatives (FN) for English tasks compared to the neutral condition in percent}
    \label{tab:change}
    \begin{tabular}{l l r r}\hline
        \textbf{Model} & \textbf{Condition} & \textbf{$\Delta$ TN} & \textbf{$\Delta$ FN}\\\hline
        Llama-3.1-8B & agree & 68.25 & -28.91\\
Llama-3.1-8B & disagree & 34.28 & 49.74\\
Llama-3.1-8B & negated & 55.20 & -33.33\\
Llama-3.1-8B & yesno & 91.69 & 12.11\\
Mistral-Small-24B & agree & 181.20 & 42.12\\
Mistral-Small-24B & disagree & -15.96 & -51.34\\
Mistral-Small-24B & negated & 134.10 & -5.85\\
Mistral-Small-24B & yesno & 180.80 & 48.92\\
gemma-2-27b-it & agree & 71.02 & 231.34\\
gemma-2-27b-it & disagree & 141.21 & 1527.19\\
gemma-2-27b-it & negated & 65.78 & 114.29\\
gemma-2-27b-it & yesno & 116.84 & 330.41\\
Llama-3.3-70B & agree & 6.92 & 39.62\\
Llama-3.3-70B & disagree & -81.43 & -79.97\\
Llama-3.3-70B & negated & 1.12 & 16.89\\
Llama-3.3-70B & yesno & 20.12 & 79.68\\
gpt-4o-2024-08-06 & agree & 37.02 & -9.26\\
gpt-4o-2024-08-06 & disagree & -52.61 & 21.09\\
gpt-4o-2024-08-06 & negated & 28.79 & -23.35\\
gpt-4o-2024-08-06 & yesno & 47.98 & 28.82\\\hline
    \end{tabular}
\end{table}

%% file: table.tex
\renewcommand{\arraystretch}{0.78}

\pgfplotsinvokeforeach {Llama-3.1-8B,Mistral-Small-24B,gemma-2-27b-it,Llama-3.3-70B,gpt-4o-2024-08-06} {
\begin{table*}
    \caption{Detailed results for the #1 model}    
    \label{tab:#1}
\begin{tabular}{p{5.5cm} p{0.5cm} p{0.5cm} p{0.5cm} p{0.5cm} p{0.5cm} p{1cm} p{1cm} p{1cm}}
        \small \textbf{Condition} & \small \textbf{Pos} & \small \textbf{TP} & \small \textbf{TN} & \small \textbf{FP}& \small \textbf{FN} & \small \textbf{Accuracy} & \small \textbf{P-val} & \small \textbf{Chi}\\
        \end{tabular}
         \pgfplotsinvokeforeach {definition_classification, cuad_cap_on_liability, cuad_no-solicit_of_customers,  contract_nli_explicit_identification, hearsay, contract_nli_confidentiality_of_agreement, cuad_non-compete, agb-de,clauses-pl,overall} {
    \begin{tabular}{p{0.9\linewidth}}
    \hline
        \small \expUScore{ ##1}  \\\hline
    \end{tabular}%
   }
    
\end{table*}
}